\documentclass[10pt,journal,twocolumn,twoside]{IEEEtran}

\usepackage[utf8]{inputenc}
\usepackage{cite}
\usepackage{amsmath,amssymb,amsfonts,mathtools}
\usepackage{amsthm}
\usepackage{algorithm}
\usepackage{algpseudocode}
\usepackage{float}
\usepackage{graphicx}
\usepackage{textcomp}
\usepackage[caption=false,font=footnotesize]{subfig}
\usepackage{nicefrac}
\usepackage{booktabs}
\usepackage{multirow}
\usepackage{tabularx}
\usepackage{array}
\usepackage{url}
\usepackage{threeparttable}
\usepackage{adjustbox}
\usepackage{siunitx}
\usepackage[table]{xcolor}
\usepackage{colortbl}
\usepackage{microtype}
\usepackage{bm}
\usepackage[hidelinks]{hyperref}

\graphicspath{{figures/}}

\newtheorem{theorem}{Theorem}
\newtheorem{definition}{Definition}

\newcommand{\Method}{\textit{CACP}}
\newcommand{\method}{\textit{CoCopula}}
\definecolor{bestblue}{rgb}{0,0,0.55}

\begin{document}

\title{\textbf{CoCopula}: \textbf{Co}nformal Calibration and \textbf{Copula} Aggregation for Reliable Renewable Energy Forecasting}

\author{Alireza Moradi, Mathieu Tanneau, Reza Zandehshahvar$^{*}$, and Pascal Van Hentenryck%
\thanks{$^{*}$Corresponding author: Reza Zandehshahvar (e-mail: reza@isye.gatech.edu).}\\[4pt]
\normalsize NSF Artificial Intelligence Institute for Advances in Optimization (AI4OPT)\\
H. Milton Stewart School of Industrial and Systems Engineering\\
Georgia Institute of Technology, Atlanta, GA 30332 USA\\
\small alirezamoradi@gatech.edu, \{mathieu.tanneau, reza, pascal.vanhentenryck\}@isye.gatech.edu}

\markboth{}{Moradi \MakeLowercase{\textit{et al.}}: CoCopula: Conformal Calibration and Copula Aggregation for Reliable Renewable Energy Forecasting}

\maketitle

\begin{abstract}
The rapid growth of renewable energy penetration has intensified the need for reliable probabilistic forecasts to support grid operations at aggregated (fleet or system) levels. In practice, however, system operators often lack access to fleet-level probabilistic models and instead rely on site-level forecasts produced by heterogeneous third-party providers. Constructing coherent and calibrated fleet-level probabilistic forecasts from such inputs remains challenging due to complex cross-site dependencies and aggregation-induced miscalibration. Motivated by a real-world operational challenge identified through collaboration with an industry partner, this paper presents a probabilistic aggregation and calibration framework that directly converts site-level probabilistic forecasts into reliable fleet-level forecasts. The paper introduces \method~ (\textbf{Co}nformal Calibration and \textbf{Copula} Aggregation), which integrates Copula-based dependence modeling to capture cross-site correlations followed by Context-Aware Conformal Prediction (CACP) to  mitigate miscalibration at the aggregated level. This combination enables dependence-aware aggregation while  improving empirical coverage and maintaining sharp prediction intervals. Experiments on large-scale solar generation datasets from three system operators in the United States including MISO, ERCOT, and SPP demonstrate that the proposed \method{} approach achieves near-nominal empirical coverage and consistently lower Winkler scores than competing aggregation-based baselines across different target coverage levels. Importantly, \method{} achieves performance comparable to directly calibrated system-level forecasts, demonstrating its practical value when dedicated fleet-level forecasting models are unavailable.
\end{abstract}


\section{Introduction}
\label{sec:intro}

\begin{figure*}[t!]
    \centering
    \includegraphics[width=1\linewidth]{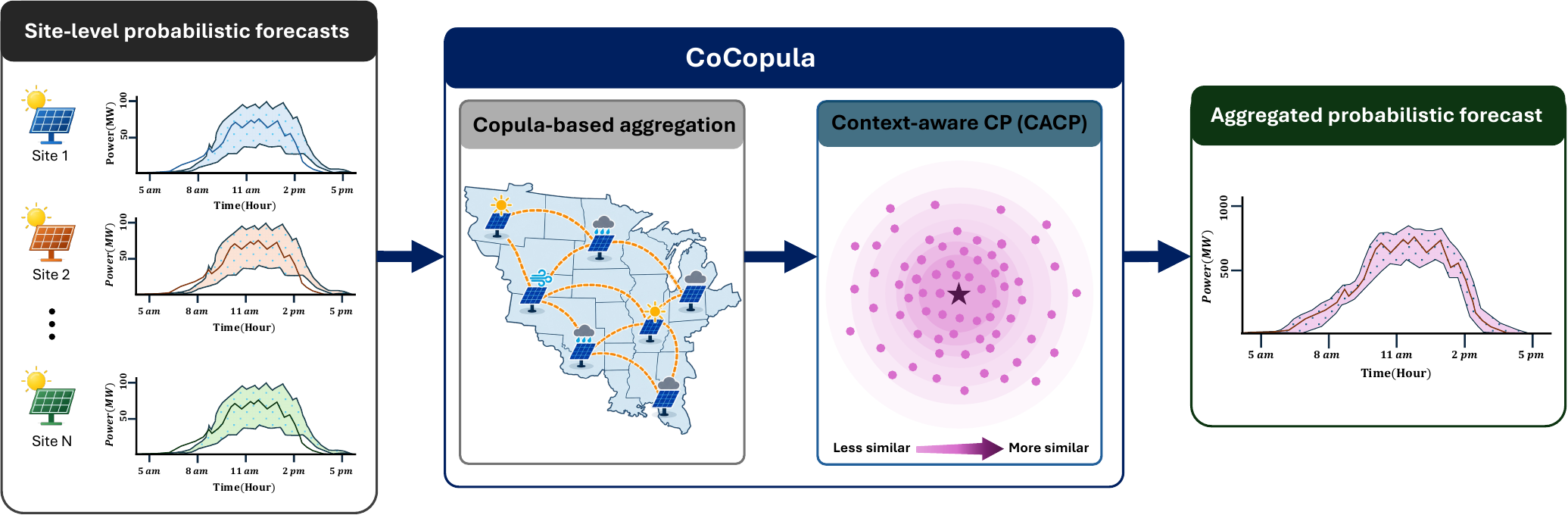}
    \caption{Conceptual overview of the proposed \method{} framework. Site-level probabilistic forecasts are first combined using copula-based aggregation to capture cross-site dependencies and construct a fleet-level predictive distribution. Context-aware conformal prediction (CACP) then calibrates the aggregated prediction intervals using historical observations and contextual similarity, producing reliable and sharp fleet-level forecasts without requiring a dedicated system-level forecasting model.}

    \label{fig:figure1}
\end{figure*}

\subsection{Motivation and Related Works}

As the penetration of renewable energy sources continues to increase in modern power systems, operational and planning uncertainty has grown substantially. This transition toward weather-dependent generation introduces significant variability, making deterministic forecasts insufficient for reliable grid operation. As a result, probabilistic forecasting has become a critical component of decision-making in power systems by providing uncertainty-aware estimates of generation, demand, and other key system variables \cite{teixeira2024advancing}.

Motivated by this need, considerable advances have been made in probabilistic renewable energy forecasting. Existing methods span a broad methodological spectrum, ranging from classical statistical approaches such as ARIMA \cite{atique2019forecasting}, to machine learning techniques including Random Forests \cite{shi2018improved} and Support Vector Machines \cite{zeng2013short}, and more recently to deep learning architectures such as recurrent neural networks (RNNs) \cite{xia2021stacked} and other advanced neural forecasting models \cite{jonkers2024novel}. Comprehensive reviews of probabilistic forecasting methods for renewable energy can be found in \cite{sweeney2020future,zhang2014review,hong2020energy,wang2021review}; see also \cite{hong2016probabilistic} for the Global Energy Forecasting Competition 2014 and \cite{wan2013probabilistic} for an early probabilistic wind power forecasting method based on extreme learning machines.

Despite these advances, constructing accurate probabilistic forecasts at high aggregation levels, such as regional zones or system-wide renewable fleets, remains a significant challenge. Developing and maintaining dedicated fleet-level forecasting models may require high-resolution meteorological inputs, site-specific operational characteristics, and continuously updated configuration information across a large number of assets. Collecting, maintaining, and synchronizing such data at scale is often costly and operationally demanding. Furthermore, aggregated renewable generation exhibits complex spatio-temporal dependencies arising from heterogeneous site behaviors and large-scale weather dynamics, which further complicate the development and maintenance of such models \cite{verdone2025review,benti2023forecasting}.

In many practical settings, probabilistic forecasts are available at lower hierarchical levels (e.g., individual sites or sub-regions) but not at higher aggregation levels (e.g., fleet or system level), despite their critical importance for grid operation and optimization. This has led to growing interest in probabilistic forecast aggregation within hierarchical renewable energy systems. Existing approaches include copula-based bottom-up frameworks that construct coherent probabilistic forecasts across the levels of a hierarchy, developed for electricity demand \cite{taieb2017coherent,taieb2021hierarchical}, copula-based conditional aggregation methods that model the dependence between individual wind farms and their aggregated output \cite{sun2019conditional}, and copula models that explicitly learn cross-site dependence structures for fleet-level forecasting \cite{moradi2025copula,zhang2025weather}. Copulas have also been used at the single-site level, e.g., to model the joint distribution of solar power and ambient temperature within a Bayesian forecasting framework \cite{panamtash2020copula}. However, the aggregated forecasts produced by these methods may remain miscalibrated, primarily due to imperfect dependence modeling, structural errors in site-level probabilistic forecasts, and model misspecification. Therefore, accounting for cross-site dependencies alone does not necessarily ensure reliable uncertainty estimates at the aggregated level, motivating the need for post-aggregation calibration.

Conformal prediction (CP) provides a principled approach to address the miscalibration of probabilistic forecasts, offering finite-sample marginal coverage guarantees under exchangeability in its standard form while maintaining relatively low computational complexity \cite{vovk2005algorithmic}. Its adoption in renewable energy forecasting has grown rapidly, particularly for uncertainty quantification in solar and wind power generation. For instance, \cite{jonkers2024novel} combines split CP with quantile random forests for wind power forecasting, while \cite{wang2023conformal} proposes an asymmetric multi-quantile conformal adjustment strategy to improve day-ahead wind prediction intervals. In solar generation forecasting, \cite{renkema2024enhancing} evaluates weighted, KNN-based, and Mondrian CP variants for constructing calibrated intervals. More recently, \cite{moradi2025enhanced} introduces context-aware conformal prediction (CACP), which incorporates contextual weighting to produce efficient (i.e., sharper) prediction intervals tailored to renewable energy applications.

Motivated by an operational need identified through collaboration with an industry partner, this paper presents \method{}, a probabilistic aggregation and calibration framework for settings in which site-level probabilistic forecasts are available but dedicated fleet-level forecasting models are not. \method{} uses copula-based dependence modeling to construct a fleet-level predictive distribution from site-level forecasts. The resulting outputs are then calibrated using CP. Overall, \method{} enables calibrated fleet-level probabilistic forecasts without the need to retrain site-level forecasting models or accessing their proprietary model inputs, while relying on historical observations for dependence estimation and calibration.

\subsection{Contributions and Outline}
This paper makes the following key contributions:
\begin{itemize}
    \item Proposes a calibrated aggregation framework that takes probabilistic forecasts from arbitrary site-level forecasting models as input and produces reliable and sharp fleet-level probabilistic forecasts by combining copula-based aggregation with context-aware conformal prediction.
    \item Addresses a practical industry challenge in renewable energy forecasting, where reliable fleet- or system-level probabilistic forecasts are often unavailable despite the availability of forecasts at individual site or zone levels.
    \item Demonstrates the effectiveness of \method{} through extensive experiments on large U.S. power systems, including the Midcontinent Independent System Operator (MISO), the Southwest Power Pool (SPP), and the Electric Reliability Council of Texas (ERCOT). The proposed method is compared against several baselines, including the system-level probabilistic forecasts of the National Renewable Energy Laboratory (NREL) \cite{NREL_PERFORMDataset}, independent aggregation, uncalibrated copula aggregation, and conformal calibration approaches applied at both the system and aggregated levels. Results show that \method{} consistently achieves near-target coverage and the lowest Winkler scores among the aggregation-based methods at all target levels, with performance comparable to directly calibrated system-level forecasts.

\end{itemize}

The remainder of the paper is organized as follows. Section~\ref{sec:problem} formulates the problem, Section~\ref{sec:method} presents the methodology, Section~\ref{sec:exp} describes the experimental setting, Section~\ref{sec:results} reports the numerical results, and Section~\ref{sec:conclusion} concludes the paper.

\section{Problem Formulation}
\label{sec:problem}

This section formalizes the probabilistic aggregation setting considered in this work. Marginal probabilistic forecasts are assumed to be available at the individual site level. The objective is to construct a reliable fleet-level probabilistic forecast that (i) coherently accounts for cross-site dependence structures and (ii) achieves calibrated uncertainty coverage at the aggregated level.

\subsection{Problem Setting}
\label{sec:problem:setting}

Consider a system comprising $\mathcal{N}=\{1,\dots,N\}$ renewable generation sites (e.g., solar plants) observed over historical time indices $\mathcal{T}=\{1,\dots,T\}$.  
Let $x_{i,t}$ denote the realized power output of site $i\in\mathcal{N}$ at time $t\in\mathcal{T}$, and define the matrix of all observations as:

\begin{align}
\mathbf{X}_{.,\mathcal{T}}=
    \begin{bmatrix}
        \vert & \dots & \vert \\
        \mathbf{x}_{.,1} & \dots & \mathbf{x}_{.,T} \\
        \vert & \dots & \vert
    \end{bmatrix}
    \in\mathbb{R}^{N\times T},
\end{align}
where $\mathbf{x}_{.,t} = [x_{1,t}, \dots, x_{N,t}]$. For each site $i \in \mathcal{N}$ and time $t \in \mathcal{T}$, a probabilistic forecasting model provides a marginal predictive cumulative distribution function (CDF) $\hat{F}_{i,t}$. Without loss of generality, these marginal predictive distributions are assumed to be available in quantile form. The marginal forecasts are univariate and do not encode explicit spatial dependencies across sites, which commonly occurs when site-level forecasts are generated independently or obtained from heterogeneous forecasting providers.

The aggregated (fleet-level) power generation is defined as:
\begin{align}
    x_{0,t} = \sum_{i=1}^{N} x_{i,t}, \qquad t\in\mathcal{T}.
\end{align}

Given a future time index $t>T$ and site-level marginal forecasts $\{\hat{F}_{i,t}\}_{i\in\mathcal{N}}$, the objective is to construct a calibrated prediction interval for the aggregated output at a pre-specified confidence level $1-\alpha$ with $\alpha \in (0, 1)$.  
Specifically, the goal is to produce an interval
\begin{align}
    \tilde{C}^{\alpha}_{0,t}
    =
    \bigl[\tilde{q}^{\alpha/2}_{0,t},\;
          \tilde{q}^{1-\alpha/2}_{0,t}\bigr],
\end{align}
such that it satisfies the marginal coverage guarantee
\begin{align}
    \mathbb{P}\!\left(
        x_{0,t} \in \tilde{C}^{\alpha}_{0,t}
    \right)
    \ge 1-\alpha.
    \label{eq:coverage}
\end{align}

Constructing $\tilde{C}^{\alpha}_{0,t}$ is challenging for two main reasons. 
First, cross-site dependencies must be accurately captured when aggregating marginal forecasts, as ignoring spatial correlations can lead to biased or overly narrow uncertainty estimates at the fleet level. 
Second, even when site-level probabilistic forecasts are well specified, prediction intervals derived from the aggregated distribution may be miscalibrated due to error propagation and model mismatch introduced during aggregation.

To address these challenges, Section~\ref{sec:method} outlines the proposed framework for constructing calibrated and sharp fleet-level prediction intervals, combining copula-based aggregation with conformal calibration techniques.

\section{Methodology}
\label{sec:method}

Figure~\ref{fig:figure1} provides a conceptual overview
of the proposed CoCopula framework, which consists of two
main stages. First, site-level probabilistic forecasts are
aggregated into a fleet-level predictive distribution using
a Gaussian copula, which captures cross-site dependencies
and enables probabilistic aggregation via Monte Carlo sampling.
The resulting aggregated prediction intervals are then
calibrated using \Method{} to account for context-dependent
forecast errors and improve their reliability.

\subsection{Aggregation Stage}
\label{sec:method:copula}

Copula functions provide a principled framework for constructing multivariate distributions by separating marginal behavior from the dependence structure. This property makes copulas particularly well-suited for probabilistic aggregation, where marginal forecasts are available but cross-variable dependencies must be modeled explicitly. Formally, a copula is defined as follows:

\begin{definition}[Copula]
    Let $d \in \mathbb{Z}_{+}$.
    A function $\mathbf{C}: [0, 1]^{d} \rightarrow [0, 1]$ is a \emph{copula} if it is the cumulative distribution function (CDF)
    of a $d$-dimensional random variable whose marginal distributions are all uniform on $[0,1]$.
\end{definition}

The fundamental connection between joint and marginal distributions is given by Sklar’s theorem, which underpins copula-based aggregation.

\begin{theorem}[Sklar's Theorem\cite{sklar1959fonctions}]
    Let $\mathbf{V}$ be a $d$-dimensional random variable with joint CDF $\mathbf{F}$ and marginal CDFs $(F_i)_{i=1}^{d}$.
    There exists a copula function $\mathbf{C}: [0,1]^d \rightarrow [0,1]$ such that
    \begin{align}
        \mathbf{F}(v_1, \dots, v_d) = \mathbf{C}\big(F_1(v_1), \dots, F_d(v_d)\big).
    \end{align}
\end{theorem}

Sklar’s theorem enables modeling the joint distribution of site-level generation by combining marginal predictive distributions with a copula that captures cross-site dependencies.

Following the notation introduced in Section~\ref{sec:problem}, historical observations $\{x_{i,t}\}$ and their corresponding site-level marginal predictive CDFs $\{\hat{F}_{i,t}\}$ are considered.
To capture cross-site dependencies, the probability integral transform is applied to each marginal forecast,
\begin{align}
    \hat{u}_{i,t} = \hat{F}_{i,t}(x_{i,t}),
\end{align}
where $\hat{u}_{i,t} \in [0,1]$ is uniformly distributed when the marginal forecasts are calibrated. 
Let $\Phi$ denote the standard normal CDF and define
\begin{align}
    \hat{z}_{i,t} = \Phi^{-1}(\hat{u}_{i,t}),
\end{align}
which are stacked into the matrix $\hat{\mathbf{Z}} \in \mathbb{R}^{N \times T}$. 
The empirical correlation matrix of the rows of $\hat{\mathbf{Z}}$
\begin{align}
    \hat{\boldsymbol{\Sigma}} = \operatorname{corr}(\hat{\mathbf{Z}})
\end{align}
serves as the correlation parameter of a Gaussian copula $\mathbf{C}_{\hat{\boldsymbol{\Sigma}}}$, encoding cross-site dependencies. Note that the correlation matrix is estimated using only the historical data.

Given a future time $t > T$ with available site-level marginal forecasts $\{\hat{F}_{i,t}\}_{i \in \mathcal{N}}$, the joint predictive distribution of site-level generation is represented as:
\begin{align}
    \hat{\mathbf{F}}_{t}(x_{1,t}, \dots, x_{N,t})
    =
    \mathbf{C}_{\hat{\boldsymbol{\Sigma}}}
    \big(
        \hat{F}_{1,t}(x_{1,t}), \dots, \hat{F}_{N,t}(x_{N,t})
    \big).
\end{align}

As closed-form aggregation is generally infeasible, Monte Carlo sampling is used to approximate the aggregated distribution. 
$S$ samples are drawn from a multivariate normal distribution,
\begin{align}
    \tilde{\mathbf{z}}^{(s)} \sim \mathcal{MVN}(0, \hat{\boldsymbol{\Sigma}}), \qquad s = 1, \dots, S,
\end{align}
transformed to uniforms, $\tilde{u}^{(s)}_{i} = \Phi(\tilde{z}^{(s)}_{i})$, where $\tilde{z}^{(s)}_{i}$ is the $i$-th component of $\tilde{\mathbf{z}}^{(s)}$, and mapped through the marginal inverse CDFs:
\begin{align}
    \tilde{x}^{(s)}_{i,t} = \hat{F}^{-1}_{i,t}(\tilde{u}^{(s)}_{i}),
    \qquad s = 1, \dots, S.
\end{align}
Fleet-level generation samples are obtained as follows:
\begin{align}
    \tilde{x}^{(s)}_{0,t} = \sum_{i=1}^{N} \tilde{x}^{(s)}_{i,t},
\end{align}
yielding an empirical approximation of the aggregated predictive distribution $\hat{F}_{0,t}$. Its quantiles $\hat{q}^{\beta}_{0,t} = \hat{F}^{-1}_{0,t}(\beta)$, $\beta \in (0,1)$, are the empirical quantiles of the samples $\{\tilde{x}^{(s)}_{0,t}\}_{s=1}^{S}$ and define the aggregated prediction interval $\hat{C}^{\alpha}_{0,t} = [\hat{q}^{\alpha/2}_{0,t}, \hat{q}^{1-\alpha/2}_{0,t}]$. 

\subsection{Calibration Stage}
\label{sec:method:cacp}

Although copula-based aggregation captures cross-site dependencies, prediction intervals derived from the aggregated distribution may fail to satisfy the marginal coverage guarantee in \eqref{eq:coverage}.
This motivates a post-hoc calibration step at the aggregated level.

\paragraph{Conformalized Quantile Regression (CQR)} This approach provides a distribution-free mechanism for calibrating predictive intervals by combining quantile regression with CP, yielding finite-sample marginal coverage guarantees while adapting to heteroscedastic forecast errors \cite{romano2019conformalized}.

In this section, the fleet index $0$ is dropped for readability: $y_t = x_{0,t}$ denotes the fleet-level generation, $\hat{q}^{\beta}_{t} = \hat{q}^{\beta}_{0,t}$ its aggregated quantiles, and $\tilde{C}^{\alpha}_{t} = \tilde{C}^{\alpha}_{0,t}$ the calibrated interval. The index $t$ refers to a test hour and $\tau$ to a calibration hour.
Given a miscoverage level $\alpha \in (0,1)$, CQR starts from the prediction interval $\hat{C}^{\alpha}_{t} = [\hat{q}^{\alpha/2}_t, \hat{q}^{1-\alpha/2}_t]$.
Calibration is performed on a held-out calibration set $\mathcal{T}_{cal}$ by defining, for each $\tau \in \mathcal{T}_{cal}$, the conformity score
\begin{align}
    s_{\tau}
    =
    \max \bigl(
        \hat{q}^{\alpha/2}_{\tau} - y_{\tau},\;
        y_{\tau} - \hat{q}^{1-\alpha/2}_{\tau}
    \bigr),
\end{align}
which measures the deviation of the observed outcome from the predicted interval.
The conformalized prediction interval is then obtained as
\begin{align}
    \tilde{C}^{\alpha}_{t}
    =
    \big[
        \hat{q}^{\alpha/2}_{t} - \hat{s},\;
        \hat{q}^{1-\alpha/2}_{t} + \hat{s}
    \big],
\end{align}
where
\begin{align}
    \hat{s} = s_{(k)}, \qquad k = \big\lceil (n+1)(1-\alpha) \big\rceil,
\end{align}
in which $s_{(1)} \le \dots \le s_{(n)}$ are the sorted calibration scores, $n = |\mathcal{T}_{cal}|$, and $\hat{s} = +\infty$ if $k > n$.
Under the exchangeability assumption, the resulting interval satisfies the desired marginal coverage guarantee \cite{romano2019conformalized}.

\paragraph{Weighted Conformal Prediction}
Standard CQR assigns equal importance to all calibration samples, implicitly assuming global exchangeability between calibration and test points.
In renewable energy forecasting, forecast errors are strongly regime-dependent, which may lead to conservative or locally miscalibrated intervals.
To address this limitation, weighted conformal prediction is adopted.

Let $t$ denote a test instance with context vector $c_t$.
Each calibration sample $\tau \in \mathcal{T}_{cal}$ is assigned a non-negative similarity weight
\begin{equation}
    w_{\tau} = \psi(c_t, c_{\tau}), \qquad w_{\tau} \ge 0,
\end{equation}
where $\psi(\cdot,\cdot)$ is a similarity function.
The weights are normalized as
\begin{equation}
    p_{\tau} = \frac{w_{\tau}}{\sum_{\tau' \in \mathcal{T}_{cal}} w_{\tau'} + w_t},
\end{equation}
where $w_t = \psi(c_t, c_t)$ is the weight of the test instance itself, which is placed as a point mass at $+\infty$ \cite{barber2023conformal}.

Instead of using an unweighted empirical quantile of conformity scores, the conformal correction is computed using a weighted quantile:
\begin{equation}
    \hat{s}
    =
    \inf\Big\{ s \in \{s_{\tau}\}_{\tau \in \mathcal{T}_{cal}} \;:\; \sum_{\tau \in \mathcal{T}_{cal}:\, s_{\tau} \le s} p_{\tau} \ge 1-\alpha \Big\},
\end{equation}
with $\hat{s} = +\infty$ if no calibration score satisfies the condition,
leading to the calibrated prediction interval
\begin{equation}
    \tilde{C}^{\alpha}_{t}
    =
    [\hat{q}^{\alpha/2}_{t} - \hat{s},\;
     \hat{q}^{1-\alpha/2}_{t} + \hat{s} ].
\end{equation}

When the weights are uniform, the procedure reduces to standard CQR. When the data are not exchangeable, \cite{barber2023conformal} bound the loss of coverage relative to $1-\alpha$ by a weighted measure of the deviation from exchangeability, and the guarantee is recovered when the data are exchangeable.

\paragraph{Context-Aware Conformal Prediction (CACP)}
Building on weighted CP, \Method{} implements a context-aware calibration strategy by defining similarity weights using physically meaningful auxiliary features, enabling calibration to adapt to local operating regimes while preserving statistical validity. In this work, similarity is defined using a radial basis function (RBF) kernel, or alternatively a Laplacian kernel (below):
\begin{equation}
    \psi(c_t, c_{\tau})
    =
    \exp\!\left(-\gamma \lVert c_t - c_{\tau} \rVert^2 \right),
\end{equation}
where $\gamma > 0$ controls the degree of localization of the weighting.
This kernel-based similarity emphasizes calibration samples that are most relevant to the test instance.
Although the framework allows any non-negative similarity function, two kernels are considered in this study: the RBF kernel above and the Laplacian kernel $\psi(c_t, c_{\tau}) = \exp(-\gamma \lVert c_t - c_{\tau} \rVert_1)$. For each test window, the kernel type and $\gamma \in \{0.5, 1, 2\}$ are selected on the validation week by minimizing the Winkler score averaged over the target coverage levels, and the context features are min--max scaled before distances are computed.

The context vector $c_t$ is constructed from physically meaningful auxiliary features designed to capture temporal structure and operating regimes in solar generation.
Specifically, it includes lagged historical generation values ($H_t \in \mathbb{R}^{24}$, the realized aggregate generation at each of the 24 hours of the previous day), periodic time embeddings ($\xi^h_t,\xi^d_t,\xi^m_t$, sine and cosine encodings of the hour of day, day of year, and month), and a normalized representation of the solar day ($\xi^s_t$).
Formally,
\begin{equation}
    c_t =
    \left(
    H_t,
    \xi^h_t,
    \xi^d_t,
    \xi^m_t,
    \xi^s_t
    \right).
\end{equation}

The full context vector is used in every test window. These features allow \Method{} to adapt prediction intervals to local operating conditions while maintaining statistical validity.

\section{Experimental Setting}
\label{sec:exp}
\subsection{Data}
\begin{figure}
    \centering
    \includegraphics[width=0.9 \linewidth]{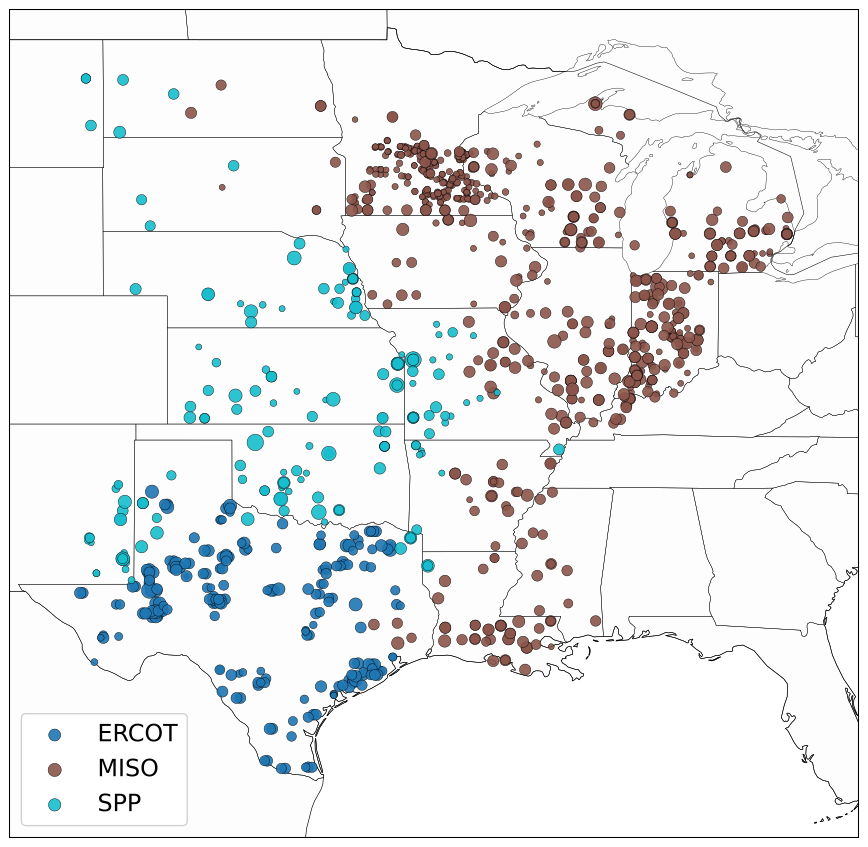}
    \caption{Spatial distribution of solar generation sites across MISO, ERCOT, and SPP, illustrating the geographic diversity of the systems evaluated in this work.}
    \label{fig:map}
\end{figure}

The experiments are performed on large-scale datasets from key U.S. power markets, including MISO, SPP, and ERCOT. The evaluation relies on day-ahead solar generation forecasts, obtained from NREL \cite{NREL_PERFORMDataset}. The dataset contains hourly actual measurements and quantile-based forecasts for 2019 (MISO and SPP) and 2018 (ERCOT), covering 1149 solar sites in total (751 in MISO, 172 in SPP, and 226 in ERCOT), whose locations are shown in Figure~\ref{fig:map}. Night-time hours with zero realized generation are excluded from calibration and evaluation.

The first six months of observations serve as the initial calibration period and for learning cross-site correlations. The test period (July--December) is divided into consecutive 7-day windows, and aggregated forecasts are recalibrated weekly, once per window, while correlation structures are updated monthly. The most recent week of data before each test window is reserved for validation and hyperparameter selection (in the case of CACP), and calibration is performed using all information available prior to the test window. The hyperparameters are summarized in Table~\ref{tab:hyperparams} in the Appendix.

\begin{figure*}[ht!]
    \centering
    \includegraphics[width=\textwidth]{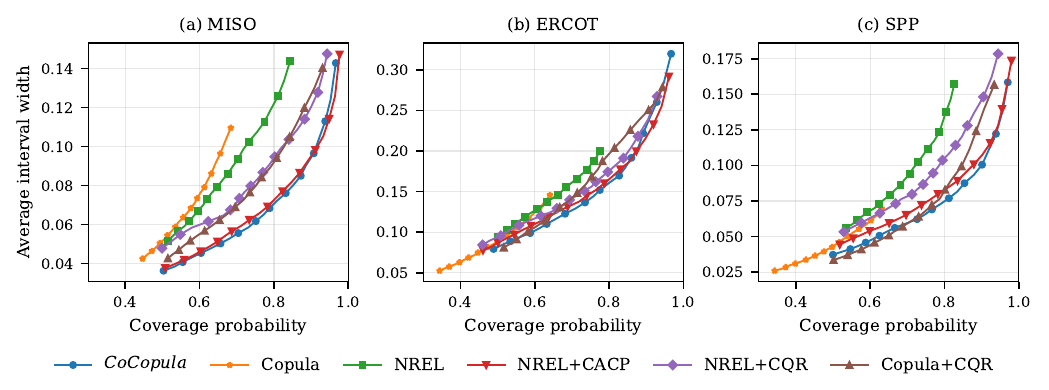}
    \caption{Coverage–sharpness trade-off for the MISO, ERCOT, and SPP systems. The figure shows the relationship between empirical coverage and average interval width for different aggregation and calibration strategies, for target coverage levels from 50\% to 95\%. \method{} achieves the lowest interval width among the aggregation-based methods for a given coverage and closely tracks NREL+CACP.}
    \label{fig:picp_aiw_miso}
\end{figure*}

\subsection{Baselines}
The proposed method is evaluated against the following baseline approaches:
\begin{itemize}
    \item \textbf{NREL:} Raw fleet-level solar power forecasts provided by NREL in the form of 99 predictive quantiles. For each target coverage level, the corresponding quantiles are selected to construct prediction intervals.
    \item \textbf{NREL+CQR:} System-level forecasts from NREL calibrated using conformalized quantile regression (CQR).
    \item \textbf{NREL+CACP:} System-level forecasts from NREL calibrated using Context-Aware Conformal Prediction (CACP).
\end{itemize}
The three NREL-based methods rely on a dedicated system-level forecast, which is assumed to be not available in the setting studied here. They are included only as references, to show the performance attainable when direct system-wide forecasts are available.
\begin{itemize}
    \item \textbf{Ind:} Fleet-level probabilistic forecasts obtained by aggregating site-level forecasts under independence ($\hat{\boldsymbol{\Sigma}} = I$), i.e., without modeling cross-site dependence.
    \item \textbf{Copula:} Fleet-level probabilistic forecasts obtained by learning cross-site dependencies via a copula-based aggregation of site-level forecasts.
    \item \textbf{Copula+CQR:} Copula-based fleet-level forecasts further calibrated using conformalized quantile regression.
\end{itemize}
\subsection{Evaluation metrics}
\label{sec:metrics}
The methods are evaluated using standard performance metrics commonly adopted in conformal prediction to assess both reliability and sharpness. Let $\mathcal{T}_{\text{test}}$ denote the set of time indices in a test window, with cardinality $|\mathcal{T}_{\text{test}}|$. For each time step $t \in \mathcal{T}_{\text{test}}$, let $y_t$ denote the observed value and let $\hat{C}_t^{\alpha} = [\hat{C}_t^{\alpha,l}, \hat{C}_t^{\alpha,u}]$ represent the $(1-\alpha)$ prediction interval produced by a given method, calibrated or not. The evaluation metrics are defined as follows:

\paragraph{Prediction Interval Coverage Probability (PICP)}
This metric measures the empirical coverage rate of the prediction intervals over the test horizon and is defined as:
\begin{align}
    \label{eq:picp}
    \text{PICP}^{\alpha} = \frac{1}{|\mathcal{T}_{\text{test}}|}\sum_{t \in \mathcal{T}_{\text{test}}} \mathbf{1}\{y_t \in \hat{C}_t^{\alpha}\}.
\end{align}

\paragraph{Average Interval Width (AIW)}
This metric evaluates the sharpness of the prediction intervals at a target coverage level $1-\alpha$ by averaging the interval widths across the test period:
\begin{align}
    \text{AIW}^{\alpha} = \frac{1}{|\mathcal{T}_{\text{test}}|} \sum_{t \in \mathcal{T}_{\text{test}}} \text{IW}^{\alpha}_t, \\
    \text{IW}^{\alpha}_{t} = \hat{C}_t^{\alpha,u} - \hat{C}_t^{\alpha,l}.
    \label{eq:aiw}
\end{align}

\paragraph{Winkler Score (WS)}
The Winkler Score provides a combined assessment of interval sharpness and coverage by penalizing prediction intervals that fail to contain the observed value. It is computed by averaging the per-time-step scores over the test horizon:
\begin{equation}
    \text{WS}^{\alpha} = \frac{1}{|\mathcal{T}_{\text{test}}|}\sum_{t \in \mathcal{T}_{\text{test}}} \text{WS}_t^{\alpha},
\end{equation}
where
\begin{align}
    \text{WS}_t^{\alpha} =
    \begin{cases}
        \text{IW}_t^\alpha+\tfrac{2}{\alpha}(y_t-\hat{C}_t^{\alpha,u}), & y_t > \hat{C}_t^{\alpha,u}, \\[6pt]
        \text{IW}_t^\alpha+\tfrac{2}{\alpha}(\hat{C}_t^{\alpha,l}-y_t), & y_t < \hat{C}_t^{\alpha,l}, \\[6pt]
        \text{IW}_t^\alpha, & \text{otherwise}.
    \end{cases}
    \label{eq:ws}
\end{align}

All time series are normalized by the maximum generation capacity of their corresponding sites or systems to ensure comparability across different aggregation levels. Each metric is computed separately within every weekly test window, and the reported values are the averages over all test windows, each window weighted equally.

\section{Numerical Results}
\label{sec:results}
\subsection{Main Analysis}
This section evaluates the performance of \method{} across three large U.S. power systems: MISO, ERCOT, and SPP. The evaluation focuses on both coverage and sharpness, using evaluation metrics discussed in Section~\ref{sec:metrics}. The primary objective is to investigate whether reliable and sharp fleet-level forecasts can be constructed solely from site-level predictions, achieving performance comparable to that obtained when dedicated fleet-level forecasts are available. Accordingly, NREL system-level forecasts are included as reference benchmarks rather than methods the proposed framework necessarily aims to outperform.

\begin{figure*}[!t]
    \centering
    \includegraphics[width=\textwidth]{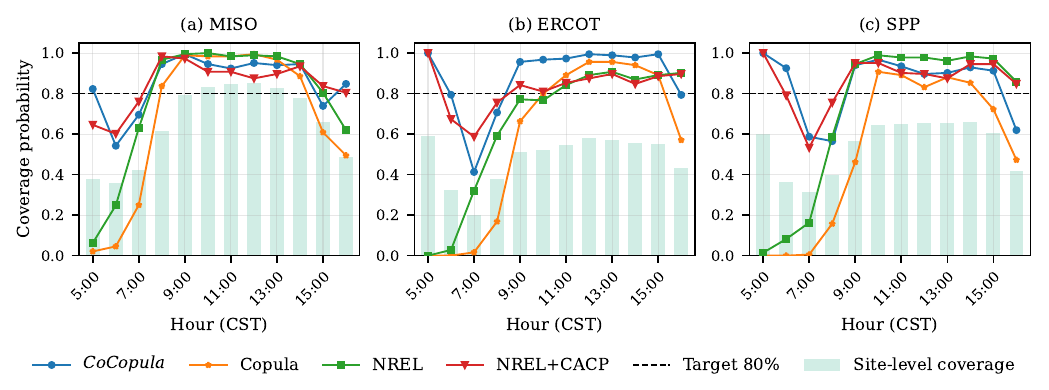}
    \caption{Hourly empirical coverage for the MISO, ERCOT, and SPP systems at target coverage $1-\alpha=80\%$. Solid lines denote system-level coverage for different methods, while the semi-transparent bars indicate the corresponding site-level coverage.}
    \label{fig:hourly_miso}
\end{figure*}

\begin{table*}[t!]
\centering

\caption{\small Fleet-level forecasting performance across MISO, ERCOT, and SPP at four target coverage levels, averaged over weekly test windows. PICP, AIW, and WS denote empirical coverage, average interval width, and Winkler score, respectively. Lowest WS among the aggregation-based methods in \textcolor{bestblue}{\textbf{blue}}, second lowest underlined; $^\dagger$ NREL-based system-level forecasts are included as reference benchmarks and excluded from ranking.}

\label{tab:main_results}\label{tab:iso_summary_combined}\label{tab:results:iso_90}
\resizebox{\textwidth}{!}{%
\scriptsize
\begin{tabular}{llrrrrrrrrrrrr}
\toprule
\textbf{ISO} & \textbf{Method}
& \multicolumn{3}{c}{$(1{-}\alpha)=90\%$}
& \multicolumn{3}{c}{$(1{-}\alpha)=80\%$}
& \multicolumn{3}{c}{$(1{-}\alpha)=70\%$}
& \multicolumn{3}{c}{$(1{-}\alpha)=60\%$} \\
\cmidrule(lr){3-5}
\cmidrule(lr){6-8}
\cmidrule(lr){9-11}
\cmidrule(lr){12-14}
& & PICP $\uparrow$ & AIW $\downarrow$ & WS $\downarrow$
& PICP $\uparrow$ & AIW $\downarrow$ & WS $\downarrow$
& PICP $\uparrow$ & AIW $\downarrow$ & WS $\downarrow$
& PICP $\uparrow$ & AIW $\downarrow$ & WS $\downarrow$ \\
\midrule

\multirow{7}{*}{ERCOT}
& NREL$^\dagger$
& 0.7501 & 0.1853 & 0.3488
& 0.6910 & 0.1585 & 0.2529
& 0.6249 & 0.1350 & 0.2107
& 0.5541 & 0.1132 & 0.1826 \\

& NREL+CQR$^\dagger$
& 0.8760 & 0.2165 & 0.3230
& 0.7736 & 0.1660 & 0.2479
& 0.6864 & 0.1379 & 0.2079
& 0.5797 & 0.1124 & 0.1806 \\

& NREL+CACP$^\dagger$
& 0.9209 & 0.2317 & 0.2878
& 0.8080 & 0.1678 & 0.2305
& 0.6863 & 0.1308 & 0.1977
& 0.5662 & 0.1024 & 0.1735 \\

\cmidrule(lr){2-14}
& Ind
& 0.1449 & 0.0194 & 0.9719
& 0.1198 & 0.0151 & 0.5095
& 0.0926 & 0.0123 & 0.3501
& 0.0746 & 0.0100 & 0.2685 \\

& Copula
& 0.6094 & 0.1301 & 0.4734
& 0.5347 & 0.1017 & 0.3280
& 0.4712 & 0.0823 & 0.2628
& 0.4056 & 0.0662 & 0.2224 \\

& Copula+CQR
& 0.9043 & 0.2510 & \underline{0.3014}
& 0.7942 & 0.1961 & \underline{0.2681}
& 0.7105 & 0.1487 & \underline{0.2388}
& 0.5990 & 0.1072 & \underline{0.2121} \\

& \method{}
& 0.9275 & 0.2620 & \textbf{{\color{bestblue}0.2924}}
& 0.8403 & 0.1804 & \textbf{{\color{bestblue}0.2283}}
& 0.7329 & 0.1375 & \textbf{{\color{bestblue}0.2021}}
& 0.6077 & 0.1052 & \textbf{{\color{bestblue}0.1876}} \\

\midrule

\multirow{7}{*}{MISO}
& NREL$^\dagger$
& 0.8085 & 0.1252 & 0.1402
& 0.7126 & 0.0977 & 0.1126
& 0.6422 & 0.0789 & 0.0958
& 0.5767 & 0.0641 & 0.0835 \\

& NREL+CQR$^\dagger$
& 0.9155 & 0.1272 & 0.1395
& 0.8145 & 0.0988 & 0.1123
& 0.7333 & 0.0795 & 0.0957
& 0.6582 & 0.0645 & 0.0835 \\

& NREL+CACP$^\dagger$
& 0.9490 & 0.1152 & 0.1212
& 0.8475 & 0.0831 & 0.0939
& 0.7336 & 0.0638 & 0.0799
& 0.6252 & 0.0499 & 0.0715 \\

\cmidrule(lr){2-14}
& Ind
& 0.1273 & 0.0100 & 0.3759
& 0.1018 & 0.0079 & 0.2000
& 0.0792 & 0.0064 & 0.1388
& 0.0638 & 0.0053 & 0.1072 \\

& Copula
& 0.6551 & 0.0958 & 0.1680
& 0.5989 & 0.0757 & 0.1260
& 0.5515 & 0.0633 & 0.1050
& 0.4946 & 0.0522 & 0.0906 \\

& Copula+CQR
& 0.8850 & 0.1194 & \underline{0.1462}
& 0.7889 & 0.0890 & \underline{0.1207}
& 0.6902 & 0.0686 & \underline{0.1038}
& 0.5840 & 0.0539 & \underline{0.0903} \\

& \method{}
& 0.9397 & 0.1130 & \textbf{{\color{bestblue}0.1234}}
& 0.8527 & 0.0806 & \textbf{{\color{bestblue}0.0974}}
& 0.7428 & 0.0622 & \textbf{{\color{bestblue}0.0852}}
& 0.6188 & 0.0483 & \textbf{{\color{bestblue}0.0768}} \\

\midrule

\multirow{7}{*}{SPP}
& NREL$^\dagger$
& 0.8085 & 0.1363 & 0.1760
& 0.7374 & 0.1060 & 0.1380
& 0.6746 & 0.0855 & 0.1161
& 0.6043 & 0.0694 & 0.1006 \\

& NREL+CQR$^\dagger$
& 0.9085 & 0.1471 & 0.1730
& 0.8104 & 0.1079 & 0.1373
& 0.7354 & 0.0860 & 0.1156
& 0.6393 & 0.0692 & 0.1002 \\

& NREL+CACP$^\dagger$
& 0.9546 & 0.1373 & 0.1455
& 0.8500 & 0.0942 & 0.1100
& 0.7322 & 0.0722 & 0.0962
& 0.6228 & 0.0569 & 0.0871 \\

\cmidrule(lr){2-14}
& Ind
& 0.2569 & 0.0189 & 0.3839
& 0.2081 & 0.0149 & 0.2116
& 0.1726 & 0.0121 & 0.1504
& 0.1393 & 0.0099 & 0.1181 \\

& Copula
& 0.5960 & 0.0608 & 0.2522
& 0.5130 & 0.0476 & 0.1620
& 0.4637 & 0.0390 & 0.1260
& 0.3999 & 0.0319 & 0.1049 \\

& Copula+CQR
& 0.8841 & 0.1239 & \underline{0.1812}
& 0.7816 & 0.0760 & \underline{0.1434}
& 0.6796 & 0.0566 & \underline{0.1186}
& 0.5812 & 0.0428 & \underline{0.1019} \\

& \method{}
& 0.9366 & 0.1211 & \textbf{{\color{bestblue}0.1366}}
& 0.8279 & 0.0816 & \textbf{{\color{bestblue}0.1087}}
& 0.7193 & 0.0621 & \textbf{{\color{bestblue}0.0978}}
& 0.6046 & 0.0480 & \textbf{{\color{bestblue}0.0900}} \\

\bottomrule
\end{tabular}
}%
\end{table*}

Table~\ref{tab:main_results} reports the numerical results for all methods across the three ISOs at four target coverage levels ($1-\alpha \in \{60\%, 70\%, 80\%, 90\%\}$). The discussion below follows the two stages of \method{}, adding one component at a time.

The simplest way to build a fleet-level forecast from site-level forecasts is to aggregate them under the assumption that site-level forecast errors are independent (Ind). This approach performs poorly: at the 90\% target level, its empirical coverage ranges from only 12.73\% to 25.69\% across the three ISOs. Since the errors of nearby sites are driven by the same weather systems, they tend to move together, and ignoring this dependence lets errors cancel out in the sum, which makes the aggregated intervals far too narrow.

Modeling the cross-site dependence with a Gaussian copula (Copula) substantially improves the aggregated forecast. At the 90\% target level, coverage increases to 65.51\%, 60.94\%, and 59.60\% in MISO, ERCOT, and SPP, respectively, and the Winkler score decreases from 0.3759 to 0.1680 in MISO, from 0.9719 to 0.4734 in ERCOT, and from 0.3839 to 0.2522 in SPP. Learning the dependence structure therefore yields considerably higher-quality fleet-level forecasts.

However, the copula-aggregated forecast still falls well short of the target coverage at every level. The aggregation combines the site-level predictive distributions as they are, so any miscalibration of these marginals is passed on to the fleet-level forecast. Figure~\ref{fig:hourly_miso} illustrates this effect: the site-level forecasts themselves under-cover, most strongly around sunrise and sunset, and the hourly coverage of the Copula forecast degrades during the same hours (see Section~\ref{sec:site_errors}). Even directly produced fleet-level forecasts are not immune to miscalibration: the NREL system-level forecasts reach only 80.85\%, 75.01\%, and 80.85\% coverage at the 90\% target in MISO, ERCOT, and SPP, respectively. A calibration step is therefore needed to make the final aggregated forecast reliable.

Calibrating the copula-aggregated forecast with CQR (Copula+CQR) brings the coverage close to the target. At the 90\% level, coverage rises to 88.50\%, 90.43\%, and 88.41\% in MISO, ERCOT, and SPP, and the Winkler score decreases to 0.1462, 0.3014, and 0.1812, respectively. CQR, however, applies a single correction computed from all calibration hours, which is justified only if calibration and test hours are exchangeable. This assumption is hard to satisfy in renewable energy forecasting, where the error distribution shifts with the season, the time of day, and the weather regime. Consistent with this, Copula+CQR remains slightly below the target coverage at all four levels in MISO and SPP.

\method{} replaces the uniform correction of CQR with CACP, which weights each calibration hour by the similarity of its context to that of the test hour, so that the correction is computed mainly from hours with comparable operating conditions. This yields both reliable coverage and sharper intervals. At the 90\% target level, \method{} achieves empirical coverage of 93.97\%, 92.75\%, and 93.66\% in MISO, ERCOT, and SPP, respectively, and it meets the target coverage at all four levels in all three ISOs. Its Winkler scores are the lowest among the aggregation-based methods at every target level, with reductions relative to Copula+CQR of 16\% in MISO, 3\% in ERCOT, and 25\% in SPP at the 90\% level. In MISO and SPP, \method{} also produces the narrowest 90\% prediction intervals among all calibrated methods.

The comparison with directly calibrated NREL system-level forecasts further demonstrates the effectiveness of \method{}. At the 90\% target level, \method{} achieves a lower Winkler score than NREL+CACP in SPP (0.1366 versus 0.1455) and slightly higher scores in ERCOT (0.2924 versus 0.2878) and MISO (0.1234 versus 0.1212). Across all ISOs and target levels, the Winkler scores of the two methods differ by at most 0.014 (8\%), with \method{} ahead in SPP at 80\% and 90\% and in ERCOT at 80\%. These findings demonstrate that reliable and sharp fleet-level probabilistic forecasts can be constructed from site-level predictions alone, achieving performance comparable to directly calibrated system-level forecasts without requiring access to dedicated fleet-level forecasting models.

Figure~\ref{fig:picp_aiw_miso} further illustrates this behavior for all three systems, showing the trade-off between empirical coverage and average interval width. \method{} achieves narrower intervals than the other aggregation-based methods at comparable coverage levels and closely tracks the directly calibrated NREL+CACP forecasts. These results further support the favorable balance between reliability and sharpness observed in Table~\ref{tab:main_results}.

Figure~\ref{fig:interval_miso} in the Appendix illustrates how the CACP step adaptively adjusts the aggregated prediction intervals on a representative day. In MISO, \method{} produces narrower intervals during peak solar generation hours, while in ERCOT and SPP, the calibration widens the intervals, most visibly near sunrise and sunset, where the uncalibrated copula forecasts fail to cover the observations. Figure~\ref{fig:width_by_hour} in the Appendix extends this analysis by comparing the mean interval widths across the entire test period. \method{} produces narrower intervals than NREL+CACP during the midday peak-generation hours in all three systems. These hours matter most for operators, since they carry the largest share of solar generation and therefore the largest absolute uncertainty to be covered by reserves. In the early-morning hours, by contrast, \method{} allows wider intervals than NREL+CACP. This shows how context-aware calibration adapts interval widths across operating conditions rather than applying a uniform correction.

\subsection{Influence of Site-Level Forecast Errors}
\label{sec:site_errors}

This section examines in more detail how calibration errors at the site level propagate to the aggregated forecasts, which motivates the calibration stage of \method{}.

Figure~\ref{fig:hourly_miso} presents the hourly empirical coverage for the MISO, ERCOT, and SPP systems at a target coverage level of $80\%$. In addition to system-level coverage curves, the figure overlays semi-transparent bars representing the corresponding site-level coverage at each hour. This visualization enables a direct comparison between site-level calibration quality and aggregated forecast performance over the diurnal cycle.

The results show that methods augmented with CACP generally exhibit more reliable hourly coverage compared to their uncalibrated counterparts. In particular, \method{} and NREL+CACP remain at or above the target coverage across most daylight hours, whereas the baseline Copula and NREL forecasts exhibit pronounced under-coverage during several early-morning and late-afternoon periods. These deviations coincide with hours characterized by low or rapidly changing solar generation, where forecasting uncertainty can vary substantially.

The site-level coverage bars reveal substantial under-coverage during sunrise and sunset hours, which in several cases coincides with the degradation observed in the aggregated Copula forecast during the same periods. This alignment suggests that miscalibration of the marginal site-level forecasts reduces the reliability of the resulting aggregated forecasts. However, inaccuracies in the estimated cross-site dependence structure may also contribute to the observed coverage errors.

Overall, the figure demonstrates that site-level forecast miscalibration can affect fleet-level uncertainty estimates even when cross-site dependencies are explicitly modeled. By calibrating the aggregated forecasts, CACP mitigates these errors and improves empirical reliability across different hours of the day, without requiring modifications to the underlying site-level forecasting models.

\section{Conclusion}
\label{sec:conclusion}

This paper proposed a probabilistic aggregation and calibration framework for renewable energy forecasting that combines copula-based dependence modeling with CACP.
The approach is motivated by a real industry setting in which reliable fleet-level probabilistic forecasts are required for system operations, while only lower-level probabilistic forecasts are available, and centralized fleet-level models cannot be trained due to data and scalability constraints.
By explicitly modeling cross-site dependencies and applying a context-aware calibration step at the aggregated level, the framework enables the construction of reliable and sharp fleet-level forecasts directly from site-level inputs.

Empirical results on large-scale solar power systems in MISO, ERCOT, and SPP show that the proposed \method{} method consistently achieves near-nominal coverage and the lowest Winkler scores among the aggregation-based methods at all target levels, with performance comparable to directly calibrated system-level forecasts.
The hourly coverage analysis further demonstrates that the proposed calibration effectively mitigates systematic miscalibration during challenging operating periods such as sunrise and sunset.

Overall, the results indicate that accurate fleet-level probabilistic forecasts can be obtained without access to raw meteorological inputs or retraining of site-level models, addressing a key limitation faced in operational practice.
Future work will explore extensions to alternative dependence structures and multi-horizon forecasting.

\section*{Acknowledgment}
This work was funded by NSF award 2112533.

\section*{{Data Availability}}

The source code and implementation of the proposed framework are publicly available at \url{https://github.com/AI4OPT/CoCopula}.

\appendix
\section{Additional Experimental Details and Results}
\label{app:additional_results}

\begin{table}[!ht]
\centering
\caption{{Hyperparameters of the copula aggregation and CACP.}}
\label{tab:hyperparams}
\footnotesize
\renewcommand{\arraystretch}{1.15}
\begin{tabular}{@{}>{\raggedright\arraybackslash}p{0.34\linewidth}>{\raggedright\arraybackslash}p{0.60\linewidth}@{}}
\toprule
\textbf{Setting} & \textbf{Value} \\
\midrule
Kernel $\psi$ & \{RBF, Laplacian\} \\
Bandwidth $\gamma$ & \{0.5, 1, 2\} \\
Validation set & Last 7 days before each test window \\
Monte Carlo samples $S$ & {1000} \\
\bottomrule
\end{tabular}
\end{table}

\begin{figure*}[!t]
    \centering
    \includegraphics[width=\textwidth]{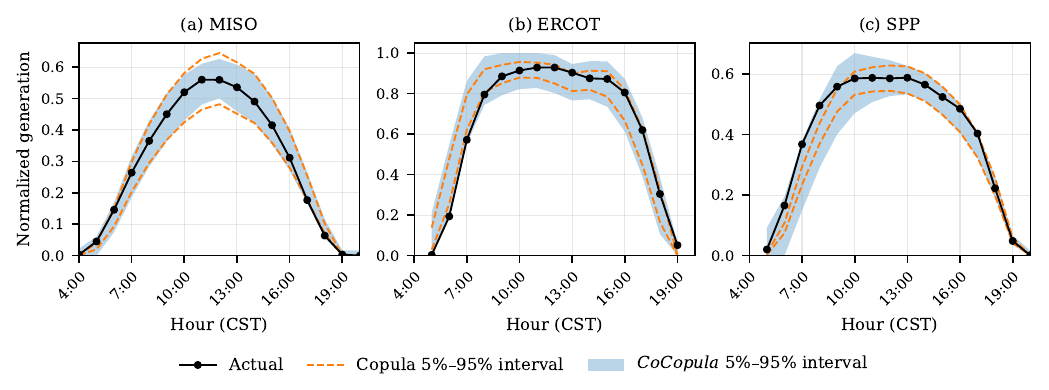}
    \caption{Example day-ahead probabilistic forecast for the MISO, ERCOT, and SPP systems on 1 July of the test year (2019 for MISO and SPP, 2018 for ERCOT). The figure compares prediction intervals produced by Copula and \method{} at the 5\%–95\% level. In MISO, the CACP calibration step notably reduces interval width during peak generation hours while maintaining coverage; in ERCOT and SPP, it widens the intervals, most visibly near sunrise and sunset, where Copula misses.}
    \label{fig:interval_miso}
\end{figure*}

\begin{figure*}[]
    \centering
    \includegraphics[width=\textwidth]{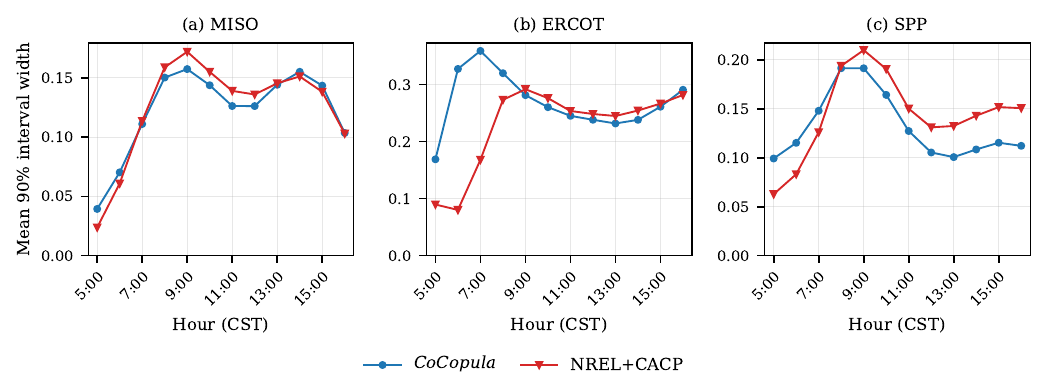}
    \caption{Mean width of the 90\% prediction interval by hour of day over the test period for MISO, ERCOT, and SPP. The figure compares \method{} with the directly calibrated NREL+CACP forecasts; \method{} gives narrower intervals during the midday peak-generation hours.}
    \label{fig:width_by_hour}
\end{figure*}

\bibliographystyle{IEEEtran}
\bibliography{references}

\end{document}